\documentclass[10pt,twocolumn,letterpaper]{article}
\usepackage[pagenumbers]{cvpr}
\usepackage{amsmath,amssymb,amsthm,mathtools}
\usepackage{graphicx,booktabs,multirow,array}
\usepackage{float}
\floatstyle{ruled}
\newfloat{algorithm}{tbp}{loa}
\floatname{algorithm}{Algorithm}
\usepackage{microtype}
\usepackage{needspace}
\usepackage{tikz}
\usetikzlibrary{arrows.meta,positioning,calc}
\definecolor{cvprblue}{rgb}{0.21,0.49,0.74}
\definecolor{rasablue}{HTML}{235A85}
\definecolor{rasaorange}{HTML}{C87530}
\definecolor{rasalight}{HTML}{EDF3F8}
\definecolor{rasagray}{HTML}{606D77}
\newcommand{\R}{\mathbb{R}}
\newcommand{\E}{\mathbb{E}}
\newcommand{\tr}{\operatorname{tr}}
\newcommand{\rank}{\operatorname{rank}}
\newcommand{\diag}{\operatorname{diag}}
\newcommand{\svd}{\operatorname{SVD}}

\newcommand{\cA}{\mathcal{A}}
\newcommand{\cD}{\mathcal{D}}
\newcommand{\cJ}{\mathcal{J}}
\newcommand{\fro}[1]{\left\|#1\right\|_{\mathrm F}}
\newcommand{\op}[1]{\left\|#1\right\|_{2}}
\DeclareMathOperator*{\argmin}{arg\,min}

\newtheorem{theorem}{Theorem}

\newtheorem{corollary}{Corollary}

\usepackage{colortbl,tabularx}
\definecolor{rasarow}{HTML}{EEF3F5}
\definecolor{tabsecondary}{HTML}{616A70}
\newcommand{\tabdelta}[1]{\mbox{\color{tabsecondary}\footnotesize $#1$}}
\newcommand{\tabdeltaavg}[1]{\mbox{\color{rasablue}\footnotesize $\mathbf{#1}$}}

\usepackage[pagebackref,breaklinks,colorlinks,allcolors=cvprblue]{hyperref}
\def\confName{CVPR}
\def\confYear{2027}
\AddToShipoutPicture{%
  \AtTextUpperLeft{%
    \put(0,22){\makebox[\textwidth][l]{\normalfont\small Preprint. Under review.}}%
    \put(0,16){\rule{\textwidth}{0.4pt}}%
  }%
}
\hypersetup{pdftitle={Same Rank, Different Needs: Rethinking LoRA Redistribution in Federated Learning},pdfauthor={Jinqian Chen, Jihua Zhu, Chang Liu}}
\title{Same Rank, Different Needs: \\ Rethinking LoRA Redistribution in Federated Learning}
\author{Jinqian Chen, Jihua Zhu\thanks{Corresponding author.}, Chang Liu\\
School of Software Engineering, Xi'an Jiaotong University\\
\texttt{chenjinqian@stu.xjtu.edu.cn}, \texttt{zhujh@xjtu.edu.cn}}
\begin{document}
\maketitle
\begin{abstract}
Federated low-rank adaptation (LoRA) enables clients with different resource budgets to collaboratively fine-tune foundation models. To distribute an aggregated update within each client’s rank budget, existing methods commonly retain its leading singular directions. However, directions that best reconstruct the update matrix need not best preserve its responses on a particular client’s inputs: a lower-energy direction may matter more to that recipient. We formalize this mismatch and show that matrix-optimal truncation can incur arbitrarily larger response error than another readout of the same rank. Motivated by this result, we propose Receiver-Aware Subspace Adaptation (RASA), a plug-in framework for receiver-aware knowledge transfer. RASA uses compact diagonal input second moments to perform uncentered PCA of receiver-weighted update responses. The resulting readout is globally optimal for the rank-constrained receiver objective and preserves at least as much response energy as spectral truncation under the diagonal metric. RASA also introduces spectral-reference aggregation, which removes the personalized initialization offset while retaining the locally learned increment. Experiments on DomainNet-100, NICO++, GLUE, and SuperNI demonstrate RASA’s effectiveness across methods under resource, task, and data heterogeneity. In particular, RASA improves DomainNet-100 accuracy by 3.23 percentage points with FlexLoRA and achieves gains of up to 8.33 ROUGE-L points on seen SuperNI tasks and 4.37 on unseen tasks with modest overhead.
\end{abstract}

\section{Introduction}
Federated learning enables clients to acquire knowledge beyond their local data through collaborative model training~\cite{mcmahan2017fedavg}. For foundation models, realizing this promise requires adaptation to heterogeneous device capabilities. Low-rank adaptation (LoRA) makes local fine-tuning affordable by learning compact updates to a frozen backbone~\cite{hu2022lora}, while heterogeneous federated LoRA matches each client's adapter rank to its resource budget~\cite{cho2024hetlora,bai2024flexlora}. This supports collaboration across clients that differ in both capacity and data, including visual devices observing different categories, environments, and acquisition conditions. A central goal is to let these clients benefit from shared knowledge within the capacity they can afford.

However, acquiring knowledge at the server does not ensure that this knowledge reaches its recipients. After aggregation, the server must compress its update into a low-rank adapter for each client. A widely used solution retains the largest singular components up to the client's rank~\cite{bai2024flexlora,ning2025eclora,ha2026rblora}. This is optimal for reconstructing the update matrix and naturally accommodates different capacities. It also makes a consequential choice: directions are ranked by their global matrix energy, independently of the inputs on which a particular recipient will use them.

\emph{The strongest directions in the matrix need not be the most influential directions for the receiver.} A matrix direction contributes to a model through its interaction with the input. A large singular component can have little effect on a client whose inputs barely activate it; a smaller component can be essential for reproducing the server update on another client's data. Compression may therefore discard useful responses even when the corresponding directions are already present in the aggregate. Equal-rank clients then receive identical views of shared knowledge despite different needs. The bottleneck lies in which directions survive redistribution, and becomes especially important when only a few can be retained.

\begin{figure}[t]
\centering
\includegraphics[width=0.95\columnwidth]{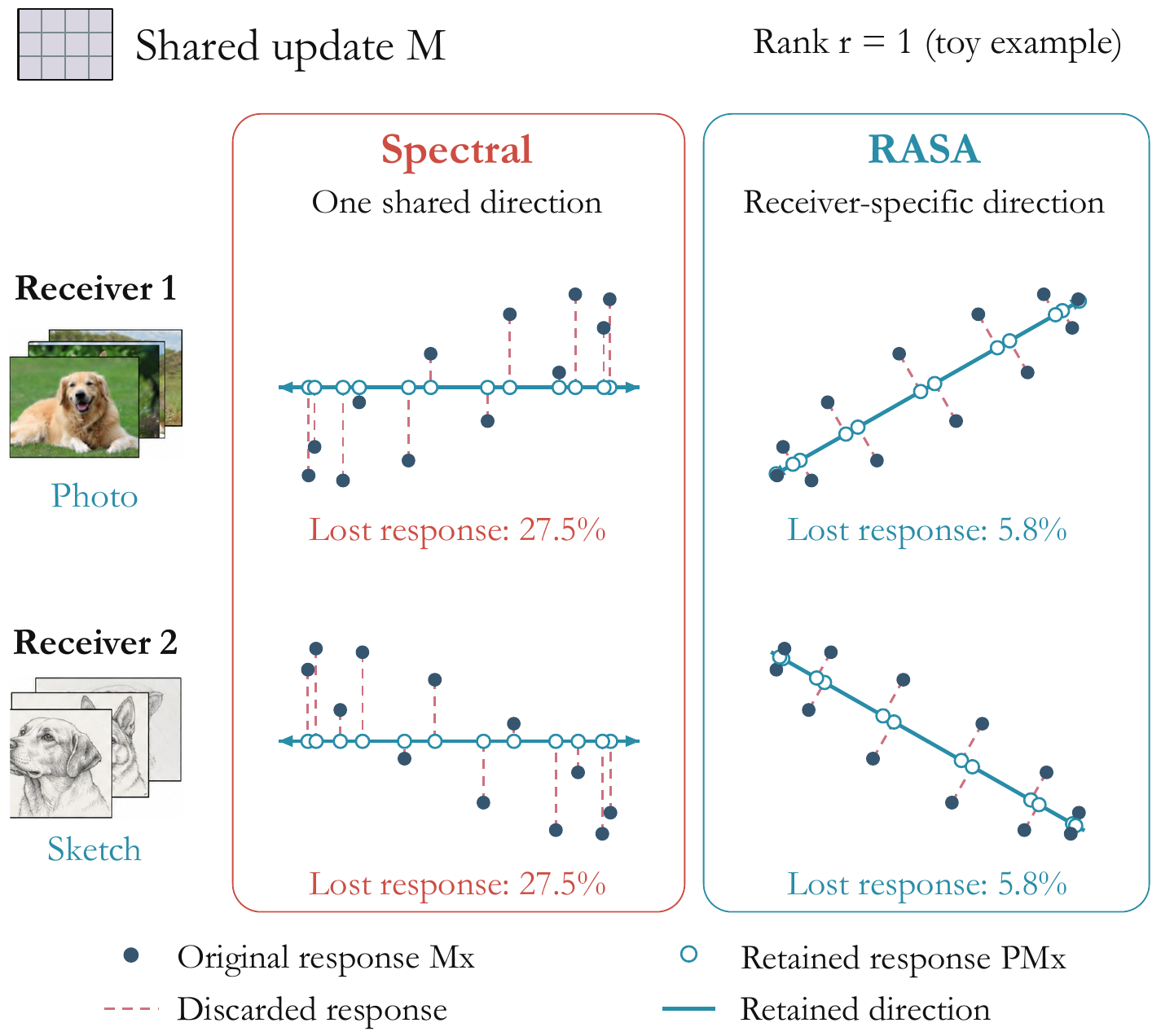}
\caption{\textbf{The same rank preserves different amounts of receiver information.} In this analytical example, dark filled dots are responses $Mx$, hollow dots are their rank-one projections $PMx$, and dashed segments show discarded responses. Spectral readout uses a common direction; RASA adapts to each receiver. Lost response is normalized squared projection error. Images indicate illustrative domains; the construction is given in the supplement.}
\label{fig:motivation}
\end{figure}

Figure~\ref{fig:motivation} contrasts these choices at the same rank-one budget. Spectral readout keeps a common matrix direction for both receivers, whereas receiver-aware readout adapts the retained direction to their response geometry. More generally, our two-direction analysis shows that the ratio between spectral and receiver-optimal response errors can be arbitrarily large. Thus, allocating sufficient capacity and allocating the right information are separate requirements. This motivates our central question: \emph{how should a server use each client's limited rank to deliver the knowledge that matters on its data?}

Our key insight is that the value of a low-rank readout should be measured through the responses it preserves for its recipient. Given server update $M$, a client with inputs $x$ should retain a matrix $C_i$ that keeps $C_ix$ close to $Mx$. This suggests selecting the principal directions of the server update's responses on that receiver. Compact diagonal input second moments estimate this geometry, while fidelity to the complete aggregate provides a global reference. The resulting uncentered principal component analysis (PCA) gives the global rank-constrained optimum in server coordinates. The rank budget determines how many directions survive; the receiver's data determine which directions they are.

Receiver-aware distribution also participates in a feedback loop. A trained adapter contains both the initialization selected for that client and the increment learned from local data. Direct aggregation carries both into the shared state, so changing the readout changes the next aggregate even when clients perform no local training. We connect personalized readout to continued collaboration through a reference-consistency principle: client-specific initializations shape local learning, while the server's spectral transition anchors the returned contributions. This gives the outgoing readout and the returning update complementary roles within one federated process.

We develop \textbf{Receiver-Aware Subspace Adaptation (RASA)} around these two linked operations. First, RASA uses compact local input statistics to extract a response-preserving adapter from the server update at the client's fixed rank. Second, RASA aggregates the locally learned increment using the corresponding spectral initialization as a reference. Clients train from their personalized readouts, while the server removes their direct distribution offsets on return. This allows personalization to shape what clients learn and pass back to the federation. RASA operates as a plug-in to SVD-based methods, retaining ordinary LoRA training and the assigned client ranks.

Our contributions are summarized as follows:
\begin{itemize}[leftmargin=*,itemsep=2pt,topsep=3pt]
\item We identify receiver-response loss as a redistribution bottleneck, showing why matrix-optimal truncation can discard client-relevant directions.
\item We propose RASA, coupling compact response PCA with spectral-reference aggregation. We prove readout optimality under the diagonal receiver metric and reference consistency without local learning, and analyze sensitivity to estimated statistics.
\item We conduct extensive experiments on visual and language benchmarks, demonstrating the effectiveness of RASA across diverse baselines under data, resource, and task heterogeneity.
\end{itemize}

\section{Related Work}
\label{sec:related}
\noindent\textbf{Resource adaptation and aggregation.}
HetLoRA uses heterogeneous ranks and local pruning~\cite{cho2024hetlora}, while FlexLoRA redistributes aggregated updates by spectral truncation~\cite{bai2024flexlora}. FLoRA stacks factors to preserve update products~\cite{wang2024flora}; FedEx-LoRA corrects factor-aggregation error~\cite{singhal2025fedex}. ECLoRA combines randomized decomposition with error compensation~\cite{ning2025eclora}, and RB-LoRA balances rank-wise contributions~\cite{ha2026rblora}. LoRA-A$^2$ alternates factor freezing and adapts ranks~\cite{koo2025loraa2}; LoRA-FAIR jointly refines aggregation and initialization~\cite{bian2025lorafair}. RASA selects receiver-dependent responses at fixed ranks and corrects the resulting initialization offset.

\noindent\textbf{Personalized federated subspaces.}
Client-specific subspaces can exploit historical update geometry~\cite{chen2026glora}, personalized spectral selection~\cite{wang2026fedrora}, activation-insensitive directions~\cite{zhang2026dysco}, or shared input-aware action geometry~\cite{du2026fedsail}. RASA studies which responses of a given server update should survive each recipient's rank budget, and couples this readout with spectral-reference aggregation. The supplement provides a mechanism-level comparison and detailed discussion.

\noindent\textbf{Activation-aware compression.}
ASVD uses activation information to rescale weights~\cite{yuan2023asvd}, and SVD-LLM uses whitening to connect spectral truncation with output reconstruction~\cite{wang2025svdllm}. RASA brings response preservation to repeated, receiver-specific redistribution under fixed client budgets. A shared aggregate is read through multiple local metrics, and the resulting initializations participate in subsequent federated optimization. Compact readout and reference correction provide a unified treatment of these two stages.

\section{What Survives Low-Rank Redistribution?}
\label{sec:problem}
The information retained by a low-rank matrix and the responses retained for a receiver are different objects. We formalize this distinction and use it to specify what a client should obtain from the server.

\noindent\textbf{Federated LoRA.}
Consider $N$ clients with local data distributions $\cD_i$ and rank budgets $r_i$. At round $t$, the active set is $\cA_t$, with normalized aggregation weights $p_i^t$. For one adapted layer, the frozen weight is $W_0\in\R^{m\times d}$ and the effective LoRA update is $s_iB_iA_i$, where $B_i\in\R^{m\times r_i}$, $A_i\in\R^{r_i\times d}$, and $s_i>0$ is the client's LoRA scale. Throughout, $M$, $C_i$, $S_i$, and $W_i$ denote \emph{effective update matrices}, including this scale and excluding $W_0$. We suppress the round and layer indices where unambiguous.

\noindent\textbf{What spectral readout optimizes.}
Let $M\in\R^{m\times d}$ be the server update at the distribution interface. By the Eckart--Young theorem~\cite{eckart1936approximation}, spectral readout returns
\begin{equation}
 S_i=\svd_{r_i}(M)
 \in\argmin_{\rank(C)\leq r_i}\fro{M-C}^2.
 \label{eq:spectral}
\end{equation}
For equal ranks and a fixed tie-breaking rule, this produces identical readouts across clients. Equation~\eqref{eq:spectral} depends on the recipient only through $r_i$.

\noindent\textbf{What the recipient experiences.}
Let $x\in\R^d$ denote a layer input under a fixed reference forward pass on local data. The error in reproducing the server update is
\begin{align}
 \mathcal{L}^{\mathrm{resp}}_i(C)
 &=\E_{x\sim\cD_i}\|(M-C)x\|_2^2\nonumber\\
 &=\tr\big[(M-C)\Sigma_i(M-C)^\top\big],
 \label{eq:response}
\end{align}
where $\Sigma_i=\E[xx^\top]$ is an \emph{uncentered} second moment. For vision transformers, $x$ is a patch or class-token representation at the adapted projection; for language models, it is a valid token representation. Holding these inputs fixed isolates the immediate effect of compression at that layer.

The metric can change the preferred subspace substantially. For $M=\diag(a,b)$, $a>b>0$, and rank one, spectral readout retains the first direction. A receiver with $\Sigma_i=\diag(\epsilon,1)$ incurs response error $b^2$. Retaining the second direction instead gives $a^2\epsilon$. For $0<\epsilon<b^2/a^2$, the latter is response-optimal, so the ratio is $b^2/(a^2\epsilon)$ and grows without bound as $\epsilon\rightarrow0$. Even an optimal matrix approximation can therefore be arbitrarily inefficient at preserving a receiver's responses under the same rank budget. The direction the receiver needs is present in $M$; it is lost at redistribution.

\noindent\textbf{A regularized receiver objective.}
To obtain a compact statistic, RASA uses $D_i=\diag(\E[x_1^2],\ldots,\E[x_d^2])$. Write $\mathcal{L}_i^D(C)=\tr[(M-C)D_i(M-C)^\top]$ for the resulting diagonal response proxy. We optimize
\begin{equation}
 \min_{\rank(C)\leq r_i}\cJ_i(C),\quad
 \cJ_i(C)=\tr\big[(M-C)G_i(M-C)^\top\big],
 \label{eq:objective}
\end{equation}
where $G_i=D_i+\lambda I_d$ and $\lambda\geq0$. The $D_i$ term prioritizes active input directions, while $\lambda\fro{M-C}^2$ preserves fidelity to the complete server update when $\lambda>0$. The setting $\lambda=0$, used in our visual experiments, retains the response objective alone. The diagonal approximation is exact when input cross moments vanish; its effect otherwise is quantified in Sec.~\ref{sec:analysis}. The target remains the complete aggregate $M$, so the feasible readout is not restricted to its leading rank-$r_i$ spectral subspace.

\section{Receiver-Aware Subspace Adaptation}
\label{sec:method}
\begin{figure*}[t]
\centering
\includegraphics[width=0.95\textwidth]{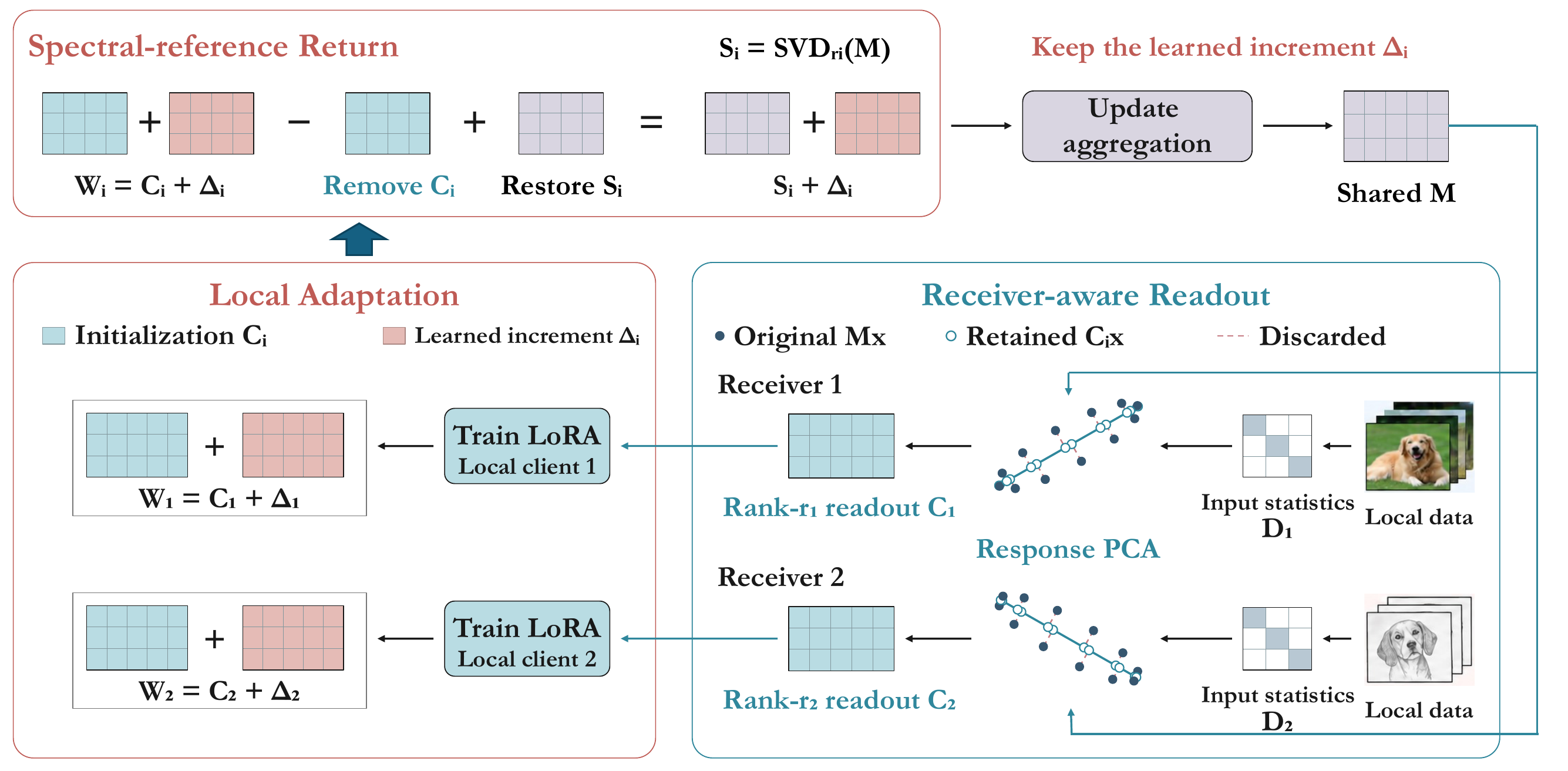}
\caption{\textbf{The federated loop of RASA.} The server distributes receiver-specific $C_i$ from shared $M$ and cached statistics $D_i$. Clients train locally and upload $W_i=C_i+\Delta_i$. The method using product averaging corrects each return to $S_i+\Delta_i$ and aggregates the corrected updates into the next $M$, preserving learned increments throughout the loop.}
\label{fig:framework}
\end{figure*}

RASA uses each client's fixed capacity to preserve the aggregate's responses on its data, then carries the resulting local learning back to the federation. These two stages are implemented by receiver-specific readout and spectral-reference aggregation, respectively. Figure~\ref{fig:framework} illustrates their interaction.

\subsection{Compact Receiver Statistics}
Each client estimates the diagonal second moment using at most 32 training examples and 4,096 valid layer-input tokens per layer. The token budget is distributed as evenly as possible across examples; valid positions are then sampled within each example, excluding padding. The estimator is
\begin{equation}
 \widehat d_i=\frac{1}{n_i^{\mathrm{tok}}}
 \sum_{x\in\mathcal{X}_i}x\odot x,\qquad
 D_i=\diag(\widehat d_i).
 \label{eq:moment}
\end{equation}
Thus each refresh transmits $d$ scalars per layer. The server caches the latest vector and its timestamp for subsequent readouts. A client without a cached statistic receives the server's spectral readout.

\subsection{PCA of Receiver Responses}
Factor the distribution target as
\begin{equation}
 M=UT,\qquad U^\top U=I_k,
 \label{eq:compact}
\end{equation}
where $U\in\R^{m\times k}$ spans the columns of $M$ and $T\in\R^{k\times d}$. The server can supply this representation directly, or it can be obtained by reduced QR from stacked low-rank factors. All receiver readouts operate on this distribution target.

An input $x$ produces the server response $Mx=U(Tx)$. Since $U$ preserves Euclidean lengths, the responses can be studied through their coordinates $Tx$. Their uncentered second moment is $T\Sigma_iT^\top$. RASA approximates this response geometry with $TD_iT^\top$ and adds the matrix-fidelity term. Define $\bar r_i=\min(r_i,k)$ and construct
\begin{equation}
 H_i=T G_iT^\top=T D_iT^\top+\lambda TT^\top.
 \label{eq:h}
\end{equation}
Let $Q_i\in\R^{k\times\bar r_i}$ contain orthonormal eigenvectors associated with its largest eigenvalues. RASA returns
\begin{equation}
 C_i=UQ_iQ_i^\top T.
 \label{eq:readout}
\end{equation}

\begin{theorem}[Optimal receiver readout]
\label{thm:optimal}
For any $G_i\succeq0$, Eq.~\eqref{eq:readout} is a global minimizer of Eq.~\eqref{eq:objective} over all matrices of rank at most $r_i$. If $\mu_1\geq\cdots\geq\mu_k\geq0$ are the eigenvalues of $H_i$, its optimal value is $\sum_{j>\bar r_i}\mu_j$.
\end{theorem}
\begin{proof}
Let $Z=MG_i^{1/2}$. Every feasible $C$ satisfies $\rank(CG_i^{1/2})\leq r_i$, so its objective is bounded below by the best rank-$r_i$ approximation error of $Z$. Since $ZZ^\top=UH_iU^\top$, the projector $P_*=UQ_iQ_i^\top U^\top$ retains a leading left singular subspace of $Z$. The choice $C=P_*M=UQ_iQ_i^\top T$ attains the bound because $CG_i^{1/2}=P_*Z$. The residual is $\sum_{j>\bar r_i}\mu_j$.
\end{proof}

RASA therefore selects the principal directions of receiver-weighted update responses. More precisely, $H_i=Y_iY_i^\top$ for $Y_i=[TD_i^{1/2},\,\sqrt{\lambda}T]$: the readout is uncentered PCA with a fidelity reference, evaluated without forming $Y_i$. The theorem needs no inverse and covers both positive fidelity regularization and the response-only setting $\lambda=0$. Even diagonal $D_i$ generally yields a dense $H_i$, allowing rotated combinations of server directions. A positive isotropic $G_i$ recovers spectral readout, and increasing $\lambda$ approaches it when the spectral boundary has an eigengap.

\subsection{Balanced LoRA Instantiation}
Write the small readout coordinates as $Q_i^\top T=L_i\Gamma_iR_i^\top$. On its nonzero singular components, set
\begin{equation}
 B_i^0=s_i^{-1/2}UQ_iL_i\Gamma_i^{1/2},\quad
 A_i^0=s_i^{-1/2}\Gamma_i^{1/2}R_i^\top.
 \label{eq:balance}
\end{equation}
Then $s_iB_i^0A_i^0=C_i$, and both factors share the singular-value scale. The client trains these ordinary LoRA factors using the client's local optimizer. If $C_i$ is zero or rank-deficient, unused directions receive a one-sided nonzero, zero-product initialization so that they remain trainable; the supplement specifies this edge case.

\subsection{Spectral-Reference Aggregation}
\label{sec:correction}
For the same server state and client rank, retain both $C_i$ and the reference $S_i=\svd_{r_i}(M)$. Local training returns $W_i=C_i+\Delta_i$. Direct averaging obeys
\begin{equation}
 \sum_i p_iW_i
 =\sum_i p_i(S_i+\Delta_i)
 +\underbrace{\sum_i p_i(C_i-S_i)}_{\text{direct distribution offset}}.
 \label{eq:offset}
\end{equation}
The last term is introduced by distribution before any client learns. RASA removes it by forming
\begin{equation}
 \widetilde W_i=W_i-C_i+S_i=S_i+\Delta_i.
 \label{eq:correction}
\end{equation}
For a method using product-space weighted averaging,
\begin{equation}
 M_{\mathrm{next}}=\sum_{i\in\cA_t}p_i^t
 \big[S_i+(W_i-C_i)\big].
 \label{eq:aggregation}
\end{equation}
The increment $\Delta_i$ is still learned from $C_i$. Therefore, personalization affects the server through the local optimization trajectory, while its direct initialization displacement is removed. The reference transition is $\sum_i p_iS_i$, which incorporates the server's rank-dependent truncation of $M$.

\noindent\textbf{Correction in shared server coordinates.}
Both initializations come from the same target: $C_i=UP_iT$ with $P_i=Q_iQ_i^\top$, and $S_i=UP_i^ST$ with $P_i^S$ the leading rank-$\bar r_i$ projector of $TT^\top$. For a spectral fallback, set $P_i=P_i^S$. Their offsets can therefore be combined before aggregation:
\begin{equation}
 \begin{aligned}
 K&=\sum_{i\in\cA_t}p_i^t(P_i^S-P_i),\\
 M_{\mathrm{next}}&=\sum_{i\in\cA_t}p_i^tW_i+UKT.
 \end{aligned}
 \label{eq:core-aggregation}
\end{equation}
This evaluates the same corrected aggregate as Eq.~\eqref{eq:aggregation} in common coordinates. The supplement gives the factorized implementation and caching requirements.

\noindent\textbf{Integration with existing methods.}
FlexLoRA applies Eq.~\eqref{eq:aggregation} before its next decomposition. For ECLoRA, the corrected incoming aggregate is passed to ECLoRA's server compression and error-compensation procedure; the same-state spectral reference is defined at its distribution interface. RB-LoRA assigns different weights to rank columns. Let $\mathcal{F}_i(B,A)=s_iB\Omega_iA$ denote its weighted contribution, with $\Omega_i$ computed from the original participating client ranks and data counts. We use
\begin{equation}
 \mathcal{F}_i(B_i^+,A_i^+)
 -\mathcal{F}_i(B_i^0,A_i^0)
 +\mathcal{F}_i(B_i^S,A_i^S).
 \label{eq:rb}
\end{equation}
All three terms use the same column weights. We store the actual dispatched factors and keep their column identities through local training. Supplementary Sec.~B gives the weights, the corresponding shared-core form, and a signed-factor representation.

\suppressfloats[t]
\begin{algorithm}[t]
\caption{RASA for a product-average aggregation}
\label{alg:rasa}
\small
\textbf{Input:} Server state, ranks $\{r_i\}$, cached moments, $\lambda\geq0$.\\
\textbf{For each communication round $t$:}
\begin{enumerate}[leftmargin=1.5em,itemsep=1pt,topsep=3pt]
\item Obtain distribution target $M^t=U^tT^t$; select $\cA_t$.
\item For each $i\in\cA_t$, set $S_i^t\gets\svd_{r_i}(M^t)$.
\item If moments are cached, form $H_i$ by Eq.~\eqref{eq:h} and $C_i^t$ by Eq.~\eqref{eq:readout}; otherwise set $C_i^t\gets S_i^t$.
\item Cache the actual dispatched initialization and reference.
\item Send factors for $C_i^t$; receive locally trained $W_i^t$.
\item Update the moment cache if a refresh is returned.
\item Form $\widetilde M^t\gets\sum_{i\in\cA_t}p_i^t(W_i^t-C_i^t+S_i^t)$, equivalently by Eq.~\eqref{eq:core-aggregation}.
\item Apply the baseline server processing to $\widetilde M^t$.
\end{enumerate}
\end{algorithm}

\begin{table*}[t]
\centering
\caption{\textbf{Visual classification with FlexLoRA.} Final-round accuracy (\%), mean $\pm$ std over three runs. ViT-B/16, $\lambda=0$; 50/30 rounds on DomainNet-100/NICO++. Avg. weights the six domains equally; shading marks RASA.}
\label{tab:vision_flex_domains}
\begingroup\small\setlength{\tabcolsep}{2pt}\renewcommand{\arraystretch}{1.10}
\begin{tabularx}{\linewidth}{@{}l*{7}{>{\centering\arraybackslash}X}@{}}
\toprule
\textbf{DomainNet-100} & Clipart & Infograph & Painting & Quickdraw & Real & Sketch & \textbf{Avg.} \\
\cmidrule(lr){2-7}\cmidrule(l){8-8}
FlexLoRA & $77.19\,{\color{tabsecondary}\scriptstyle\pm\,1.11}$ & $37.69\,{\color{tabsecondary}\scriptstyle\pm\,1.29}$ & $71.31\,{\color{tabsecondary}\scriptstyle\pm\,1.13}$ & $73.52\,{\color{tabsecondary}\scriptstyle\pm\,2.01}$ & $89.70\,{\color{tabsecondary}\scriptstyle\pm\,0.83}$ & $65.66\,{\color{tabsecondary}\scriptstyle\pm\,1.18}$ & $69.18\,{\color{tabsecondary}\scriptstyle\pm\,0.58}$ \\
\rowcolor{rasarow}
\textbf{+ RASA} & $\mathbf{80.27}\,{\color{tabsecondary}\scriptstyle\pm\,0.59}$ & $\mathbf{43.27}\,{\color{tabsecondary}\scriptstyle\pm\,1.62}$ & $\mathbf{74.71}\,{\color{tabsecondary}\scriptstyle\pm\,0.64}$ & $\mathbf{74.51}\,{\color{tabsecondary}\scriptstyle\pm\,2.31}$ & $\mathbf{91.60}\,{\color{tabsecondary}\scriptstyle\pm\,0.47}$ & $\mathbf{70.07}\,{\color{tabsecondary}\scriptstyle\pm\,0.67}$ & $\mathbf{72.40}\,{\color{tabsecondary}\scriptstyle\pm\,0.44}$ \\
{\footnotesize $\Delta$ (pp)} & \tabdelta{+3.08} & \tabdelta{+5.57} & \tabdelta{+3.40} & \tabdelta{+0.99} & \tabdelta{+1.90} & \tabdelta{+4.41} & \tabdeltaavg{+3.23} \\
\addlinespace[3pt]
\textbf{NICO++} & Autumn & Dim & Grass & Outdoor & Rock & Water & \textbf{Avg.} \\
\cmidrule(lr){2-7}\cmidrule(l){8-8}
FlexLoRA & $90.20\,{\color{tabsecondary}\scriptstyle\pm\,0.32}$ & $87.42\,{\color{tabsecondary}\scriptstyle\pm\,0.87}$ & $93.52\,{\color{tabsecondary}\scriptstyle\pm\,0.25}$ & $89.78\,{\color{tabsecondary}\scriptstyle\pm\,0.49}$ & $89.44\,{\color{tabsecondary}\scriptstyle\pm\,0.84}$ & $89.28\,{\color{tabsecondary}\scriptstyle\pm\,0.35}$ & $89.94\,{\color{tabsecondary}\scriptstyle\pm\,0.13}$ \\
\rowcolor{rasarow}
\textbf{+ RASA} & $\mathbf{90.71}\,{\color{tabsecondary}\scriptstyle\pm\,0.71}$ & $\mathbf{87.47}\,{\color{tabsecondary}\scriptstyle\pm\,1.06}$ & $\mathbf{93.84}\,{\color{tabsecondary}\scriptstyle\pm\,0.07}$ & $\mathbf{90.10}\,{\color{tabsecondary}\scriptstyle\pm\,0.28}$ & $\mathbf{90.27}\,{\color{tabsecondary}\scriptstyle\pm\,0.50}$ & $\mathbf{89.77}\,{\color{tabsecondary}\scriptstyle\pm\,0.61}$ & $\mathbf{90.36}\,{\color{tabsecondary}\scriptstyle\pm\,0.25}$ \\
{\footnotesize $\Delta$ (pp)} & \tabdelta{+0.51} & \tabdelta{+0.05} & \tabdelta{+0.33} & \tabdelta{+0.32} & \tabdelta{+0.83} & \tabdelta{+0.49} & \tabdeltaavg{+0.42} \\
\bottomrule\end{tabularx}\endgroup
\end{table*}

\begin{table*}[t]
\centering\small
\renewcommand{\arraystretch}{1.10}
\caption{\textbf{Language understanding on GLUE.} Accuracy (\%), mean $\pm$ std over three runs. MNLI-mm uses the MNLI-m checkpoint. Average weights five scores equally; $\Delta$ subtracts displayed means (pp). Bold marks the better mean.}
\label{tab:rasa-glue}
\begin{tabularx}{\linewidth}{@{}l*{6}{>{\centering\arraybackslash}X}@{}}
\toprule
Method & MNLI-m & MNLI-mm & QNLI & QQP & SST-2 & Average \\
\midrule
FlexLoRA & $64.74\,{\color{tabsecondary}\scriptstyle\pm\,1.84}$ & $66.27\,{\color{tabsecondary}\scriptstyle\pm\,1.84}$ & $82.48\,{\color{tabsecondary}\scriptstyle\pm\,0.75}$ & $78.75\,{\color{tabsecondary}\scriptstyle\pm\,0.65}$ & $90.56\,{\color{tabsecondary}\scriptstyle\pm\,0.24}$ & $76.56\,{\color{tabsecondary}\scriptstyle\pm\,0.94}$ \\
\rowcolor{rasarow}
\textbf{+ RASA} & $\mathbf{67.31}\,{\color{tabsecondary}\scriptstyle\pm\,0.50}$ & $\mathbf{69.26}\,{\color{tabsecondary}\scriptstyle\pm\,0.28}$ & $\mathbf{82.98}\,{\color{tabsecondary}\scriptstyle\pm\,0.95}$ & $\mathbf{78.89}\,{\color{tabsecondary}\scriptstyle\pm\,0.90}$ & $\mathbf{90.71}\,{\color{tabsecondary}\scriptstyle\pm\,0.35}$ & $\mathbf{77.83}\,{\color{tabsecondary}\scriptstyle\pm\,0.21}$ \\
{\footnotesize $\Delta$ (pp)} & \tabdelta{+2.57} & \tabdelta{+2.99} & \tabdelta{+0.50} & \tabdelta{+0.14} & \tabdelta{+0.15} & \tabdeltaavg{+1.27} \\
\midrule
ECLoRA & $66.60\,{\color{tabsecondary}\scriptstyle\pm\,0.19}$ & $68.87\,{\color{tabsecondary}\scriptstyle\pm\,0.41}$ & $82.83\,{\color{tabsecondary}\scriptstyle\pm\,0.57}$ & $\mathbf{79.29}\,{\color{tabsecondary}\scriptstyle\pm\,0.35}$ & $90.64\,{\color{tabsecondary}\scriptstyle\pm\,0.73}$ & $77.64\,{\color{tabsecondary}\scriptstyle\pm\,0.22}$ \\
\rowcolor{rasarow}
\textbf{+ RASA} & $\mathbf{68.41}\,{\color{tabsecondary}\scriptstyle\pm\,0.35}$ & $\mathbf{70.10}\,{\color{tabsecondary}\scriptstyle\pm\,0.35}$ & $\mathbf{83.24}\,{\color{tabsecondary}\scriptstyle\pm\,0.38}$ & $79.22\,{\color{tabsecondary}\scriptstyle\pm\,0.27}$ & $\mathbf{90.90}\,{\color{tabsecondary}\scriptstyle\pm\,0.48}$ & $\mathbf{78.38}\,{\color{tabsecondary}\scriptstyle\pm\,0.19}$ \\
{\footnotesize $\Delta$ (pp)} & \tabdelta{+1.81} & \tabdelta{+1.23} & \tabdelta{+0.41} & \tabdelta{-0.07} & \tabdelta{+0.26} & \tabdeltaavg{+0.74} \\
\midrule
RB-LoRA & $67.18\,{\color{tabsecondary}\scriptstyle\pm\,1.00}$ & $\mathbf{69.46}\,{\color{tabsecondary}\scriptstyle\pm\,0.81}$ & $82.05\,{\color{tabsecondary}\scriptstyle\pm\,1.21}$ & $78.99\,{\color{tabsecondary}\scriptstyle\pm\,0.56}$ & $90.67\,{\color{tabsecondary}\scriptstyle\pm\,0.06}$ & $77.67\,{\color{tabsecondary}\scriptstyle\pm\,0.28}$ \\
\rowcolor{rasarow}
\textbf{+ RASA} & $\mathbf{67.51}\,{\color{tabsecondary}\scriptstyle\pm\,0.59}$ & $69.39\,{\color{tabsecondary}\scriptstyle\pm\,0.71}$ & $\mathbf{82.99}\,{\color{tabsecondary}\scriptstyle\pm\,0.41}$ & $\mathbf{79.61}\,{\color{tabsecondary}\scriptstyle\pm\,0.39}$ & $\mathbf{91.06}\,{\color{tabsecondary}\scriptstyle\pm\,0.20}$ & $\mathbf{78.11}\,{\color{tabsecondary}\scriptstyle\pm\,0.34}$ \\
{\footnotesize $\Delta$ (pp)} & \tabdelta{+0.33} & \tabdelta{-0.07} & \tabdelta{+0.94} & \tabdelta{+0.62} & \tabdelta{+0.39} & \tabdeltaavg{+0.44} \\
\bottomrule\end{tabularx}
\end{table*}

\Needspace{10\baselineskip}
\section{Properties and Computational Cost}
\label{sec:analysis}

\noindent\textbf{Response preservation at the same rank.}
Optimality for the regularized objective also gives a guarantee for the unregularized response proxy $\mathcal{L}_i^D$.
\begin{corollary}[Response preservation over spectral readout]
\label{cor:response}
Let $C_i$ be the RASA readout and $S_i=\svd_{r_i}(M)$. Then
\begin{equation}
 \begin{aligned}
 \mathcal{L}_i^D(S_i)-\mathcal{L}_i^D(C_i)
 &\geq\lambda\big[\fro{M-C_i}^2-\fro{M-S_i}^2\big]\\
 &\geq0.
 \end{aligned}
 \label{eq:response-gain}
\end{equation}
\end{corollary}
\begin{proof}
Compare $C_i$ with the feasible $S_i$ in Eq.~\eqref{eq:objective}, use the Frobenius optimality of $S_i$ in Eq.~\eqref{eq:spectral}, and rearrange.
\end{proof}
Thus the fidelity term does not sacrifice response preservation relative to spectral readout under the same diagonal metric. Transfer from this proxy to the full receiver response depends on the statistic approximation considered below.

\noindent\textbf{A measurable advantage over spectral readout.}
Let $P_*=Q_iQ_i^\top$ and let $P_S$ be the rank-$r_i$ projector selected by ordinary SVD in the same server coordinates, assuming $r_i<k$. For any orthogonal projector $P$,
\begin{equation}
 \cJ_i(UPT)=\tr(H_i)-\tr(PH_i).
 \label{eq:trace}
\end{equation}
Consequently, $\cJ_i(S_i)-\cJ_i(C_i)=\tr[(P_*-P_S)H_i]\geq0$. If $\gamma_i=\mu_{r_i}-\mu_{r_i+1}>0$, then
\begin{equation}
 \cJ_i(S_i)-\cJ_i(C_i)
 \geq\frac{\gamma_i}{2}\fro{P_*-P_S}^2.
 \label{eq:gap}
\end{equation}
Together with Corollary~\ref{cor:response}, the inequality gives strict improvement in the diagonal response proxy whenever the leading subspaces differ and the receiver eigengap is positive. It supplies a layerwise diagnostic of response preservation. Section~\ref{sec:mechanism} examines response error alongside task NLL to study their empirical relationship.

\noindent\textbf{Effect of diagonal and stale statistics.}
Let $G_i^{\star}=\Sigma_i^{\mathrm{current}}+\lambda I$ describe the current full second moment, and let $\widehat G_i=\widehat D_i^{\mathrm{cached}}+\lambda I$. Let $\widehat C_i$ be the readout obtained with $\widehat G_i$, and $C_i^{\star}$ the optimum under $G_i^{\star}$. Define
\begin{equation}
 E_i=T(\widehat G_i-G_i^{\star})T^\top.
\end{equation}
The variational characterization of the leading eigenspace yields
\begin{equation}
 \cJ_i^{\star}(\widehat C_i)-\cJ_i^{\star}(C_i^{\star})
 \leq 2\bar r_i\op{E_i}.
 \label{eq:regret}
\end{equation}
Moreover, $\op{E_i}\leq\op{T}^2\op{\widehat D_i^{\mathrm{cached}}-\Sigma_i^{\mathrm{current}}}$. The latter difference separates into diagonal sampling error, temporal drift, and omitted cross moments. Thus a small upload is useful when it preserves the moment geometry relevant to $T$; increasing the token count alone cannot remove diagonal-approximation error. Proofs and this decomposition are in the supplement.

\noindent\textbf{Zero-local-update consistency.}
If clients perform no local update, $W_i=C_i$ and Eq.~\eqref{eq:correction} gives $\widetilde W_i=S_i$. Therefore the corrected incoming aggregate is exactly the baseline spectral-reference aggregate from the same state. The claim also holds under Eq.~\eqref{eq:rb}, where the trained factors equal the dispatched factors, and with identical downstream server processing. Without correction, the difference is precisely the last term in Eq.~\eqref{eq:offset}.

\noindent\textbf{Communication and computation.}
Per layer, a refreshed statistic adds $d$ scalars to the usual $r_i(m+d)$ adapter-upload scalars. If both use the same precision and $m=d$, the relative upload overhead is $1/(2r_i)$ per refresh, amortized by the refresh interval. Downlink rank and the number of locally trained adapter parameters are unchanged. After sharing $U$, $T$, and $TT^\top$ across clients, a dense construction of $H_i$ costs $O(k^2d)$ and a full eigendecomposition costs $O(k^3)$ per receiver. Instantiation costs $O(mkr_i+kr_id+r_i^2d)$. The full cost also includes obtaining the server representation and the statistic-collection forward pass.

Let $R_t=\sum_{i\in\cA_t}r_i$. Direct signed stacks use at most $3R_t$ columns. The shared-core form in Eq.~\eqref{eq:core-aggregation} admits an alternative bound of $R_t+\rank(K)\leq R_t+\min(k,2R_t)$. Both represent the same aggregate, while each client still trains and transmits rank $r_i$.

\section{Experiments}
\label{sec:experiments}
\begin{table*}[t]
\centering\small
\caption{\textbf{SuperNI after 30 federated rounds.} Free-generation ROUGE-L, mean $\pm$ std over three runs. Seen/unseen: 50/20 clients; unseen uses spectral readout. $\Delta$: paired mean gain over Pure.}
\label{tab:rasa-superni}
\begingroup\renewcommand{\arraystretch}{1.10}\setlength{\tabcolsep}{3pt}
\begin{tabularx}{\linewidth}{@{}ll*{6}{>{\centering\arraybackslash}X}@{}}
\toprule
Method & Rank profile & \multicolumn{3}{c}{In-domain (Seen)} & \multicolumn{3}{c}{Unseen task generalization} \\
\cmidrule(lr){3-5}\cmidrule(l){6-8}
& & Pure & \cellcolor{rasarow}\textbf{+ RASA} & $\Delta$ & Pure & \cellcolor{rasarow}\textbf{+ RASA} & $\Delta$ \\
\midrule
\multirow{3}{*}{FlexLoRA} & Normal & $39.62\,{\color{tabsecondary}\scriptstyle\pm\,1.38}$ & \cellcolor{rasarow}$\mathbf{44.18}\,{\color{tabsecondary}\scriptstyle\pm\,1.31}$ & \tabdelta{+4.56} & $20.20\,{\color{tabsecondary}\scriptstyle\pm\,2.59}$ & \cellcolor{rasarow}$\mathbf{22.07}\,{\color{tabsecondary}\scriptstyle\pm\,0.56}$ & \tabdelta{+1.87} \\
 & Uniform & $32.84\,{\color{tabsecondary}\scriptstyle\pm\,4.85}$ & \cellcolor{rasarow}$\mathbf{41.17}\,{\color{tabsecondary}\scriptstyle\pm\,2.03}$ & \tabdelta{+8.33} & $16.73\,{\color{tabsecondary}\scriptstyle\pm\,3.93}$ & \cellcolor{rasarow}$\mathbf{21.10}\,{\color{tabsecondary}\scriptstyle\pm\,1.40}$ & \tabdelta{+4.37} \\
 & Heavy-tail & $37.17\,{\color{tabsecondary}\scriptstyle\pm\,0.64}$ & \cellcolor{rasarow}$\mathbf{41.75}\,{\color{tabsecondary}\scriptstyle\pm\,1.21}$ & \tabdelta{+4.58} & $18.98\,{\color{tabsecondary}\scriptstyle\pm\,1.91}$ & \cellcolor{rasarow}$\mathbf{22.47}\,{\color{tabsecondary}\scriptstyle\pm\,1.24}$ & \tabdelta{+3.49} \\
\midrule
\multirow{3}{*}{ECLoRA} & Normal & $43.53\,{\color{tabsecondary}\scriptstyle\pm\,2.26}$ & \cellcolor{rasarow}$\mathbf{44.11}\,{\color{tabsecondary}\scriptstyle\pm\,2.18}$ & \tabdelta{+0.58} & $21.43\,{\color{tabsecondary}\scriptstyle\pm\,1.78}$ & \cellcolor{rasarow}$\mathbf{22.82}\,{\color{tabsecondary}\scriptstyle\pm\,1.54}$ & \tabdelta{+1.39} \\
 & Uniform & $41.93\,{\color{tabsecondary}\scriptstyle\pm\,1.23}$ & \cellcolor{rasarow}$\mathbf{42.51}\,{\color{tabsecondary}\scriptstyle\pm\,1.84}$ & \tabdelta{+0.58} & $22.48\,{\color{tabsecondary}\scriptstyle\pm\,0.75}$ & \cellcolor{rasarow}$\mathbf{22.88}\,{\color{tabsecondary}\scriptstyle\pm\,1.30}$ & \tabdelta{+0.40} \\
 & Heavy-tail & $39.82\,{\color{tabsecondary}\scriptstyle\pm\,0.93}$ & \cellcolor{rasarow}$\mathbf{40.54}\,{\color{tabsecondary}\scriptstyle\pm\,1.40}$ & \tabdelta{+0.72} & $\mathbf{23.38}\,{\color{tabsecondary}\scriptstyle\pm\,0.64}$ & \cellcolor{rasarow}$22.83\,{\color{tabsecondary}\scriptstyle\pm\,0.55}$ & \tabdelta{-0.55} \\
\midrule
\multirow{3}{*}{RB-LoRA} & Normal & $\mathbf{43.70}\,{\color{tabsecondary}\scriptstyle\pm\,2.51}$ & \cellcolor{rasarow}$43.46\,{\color{tabsecondary}\scriptstyle\pm\,2.06}$ & \tabdelta{-0.23} & $23.79\,{\color{tabsecondary}\scriptstyle\pm\,0.53}$ & \cellcolor{rasarow}$\mathbf{23.87}\,{\color{tabsecondary}\scriptstyle\pm\,1.31}$ & \tabdelta{+0.08} \\
 & Uniform & $42.94\,{\color{tabsecondary}\scriptstyle\pm\,1.54}$ & \cellcolor{rasarow}$\mathbf{43.71}\,{\color{tabsecondary}\scriptstyle\pm\,1.49}$ & \tabdelta{+0.77} & $22.53\,{\color{tabsecondary}\scriptstyle\pm\,1.00}$ & \cellcolor{rasarow}$\mathbf{23.33}\,{\color{tabsecondary}\scriptstyle\pm\,0.59}$ & \tabdelta{+0.80} \\
 & Heavy-tail & $37.13\,{\color{tabsecondary}\scriptstyle\pm\,3.75}$ & \cellcolor{rasarow}$\mathbf{40.83}\,{\color{tabsecondary}\scriptstyle\pm\,0.62}$ & \tabdelta{+3.70} & $21.80\,{\color{tabsecondary}\scriptstyle\pm\,1.91}$ & \cellcolor{rasarow}$\mathbf{22.79}\,{\color{tabsecondary}\scriptstyle\pm\,1.63}$ & \tabdelta{+0.98} \\
\bottomrule\end{tabularx}\endgroup
\end{table*}

Our comprehensive evaluation covers visual classification, language understanding, and instruction following under data heterogeneity (domains and label skew), resource heterogeneity (client ranks), and task heterogeneity (seen/unseen instructions). We test compatibility with FlexLoRA, ECLoRA, and RB-LoRA through benchmark, ablation, mechanism, sensitivity, and cost studies.

\subsection{Experimental Setup}
Our benchmarks are DomainNet-100~\cite{peng2019domainnet}, NICO++~\cite{zhang2023nicopp}, GLUE~\cite{wang2019glue}, and SuperNI~\cite{wang2022superni}. RASA preserves each method's client ranks and aggregation rule. Main visual comparisons use FlexLoRA; language and diagnostic studies also cover ECLoRA and RB-LoRA.

The main visual benchmarks use ViT-B/16~\cite{dosovitskiy2021vit}, $\lambda=0$, and 50/30 rounds on DomainNet-100/NICO++, reporting final accuracy over six domains. GLUE classification uses RoBERTa-base~\cite{liu2019roberta}, three clients, and Dirichlet $\alpha=0.1$. SuperNI instruction generation uses a 1B-parameter \mbox{LLaMA}~\cite{touvron2023llama} model, reporting round-30 ROUGE-L~\cite{lin2004rouge} on 50 seen/20 unseen clients under normal, uniform, and heavy-tail ranks; unseen tasks use spectral readout. Tables~\ref{tab:vision_flex_domains}--\ref{tab:rasa-superni} report mean $\pm$ std over seeds 42--44. Further protocols are in the supplement.

\subsection{Performance across Visual and Language Tasks}

\noindent\textbf{Visual classification.}
\label{sec:vision}
Table~\ref{tab:vision_flex_domains} shows gains across all six domains on both benchmarks. On DomainNet-100, RASA improves average accuracy from $69.18$ to $72.40$, with a paired gain of $3.23$ points. Infograph and Sketch benefit most, gaining $5.57$ and $4.41$ points. On NICO++, average accuracy increases from $89.94$ to $90.36$, with the largest gain of $0.83$ points on Rock. These results support adapting shared updates to heterogeneous visual distributions.

\noindent\textbf{Language tasks and method compatibility.}
On GLUE (Table~\ref{tab:rasa-glue}), RASA improves the averages of FlexLoRA, ECLoRA, and RB-LoRA by $1.27$, $0.74$, and $0.44$ points; thirteen of fifteen scores improve. On seen SuperNI tasks (Table~\ref{tab:rasa-superni}), it improves eight of nine comparisons. FlexLoRA gains $4.56$, $8.33$, and $4.58$ ROUGE-L under normal, uniform, and heavy-tail ranks. ECLoRA improves under every profile, and RB-LoRA gains up to $3.70$.

RASA also improves eight of nine unseen-task comparisons, including all profiles for FlexLoRA and RB-LoRA. FlexLoRA gains up to $4.37$ ROUGE-L. Since unseen tasks use spectral readout, these gains demonstrate transfer through the shared model learned during RASA training.

\subsection{Ablation and Mechanism Analysis}
\begin{table}[tb]
\centering
\caption{\textbf{Component ablation on SuperNI.} Round-30 ROUGE-L (RL, $\uparrow$) and loss ($\downarrow$), normal rank profile. The two ablated variants report single-run results; Pure baseline and Full RASA report means across runs.}
\label{tab:rasa-ablation}
\begingroup
\footnotesize
\setlength{\tabcolsep}{3pt}
\renewcommand{\arraystretch}{1.10}
\begin{tabularx}{\columnwidth}{@{}>{\raggedright\arraybackslash}Xrrrr@{}}
\toprule
Variant & Seen RL & Unseen RL & Seen loss & Unseen loss \\
\midrule
Pure baseline & 39.62 & 20.20 & 1.127 & 1.803 \\
w/o Receiver statistics & 41.43 & 21.89 & \textbf{0.9826} & \textbf{1.763} \\
w/o Update correction & 43.69 & 20.58 & 1.000 & 1.813 \\
\midrule
\rowcolor{rasarow}
\textbf{Full RASA} & \textbf{44.18} & \textbf{22.07} & 0.9929 & 1.786 \\
\bottomrule
\end{tabularx}
\endgroup
\end{table}

\begin{table*}[t]
\centering
\small
\setlength{\tabcolsep}{6pt}
\renewcommand{\arraystretch}{1.08}
\caption{\textbf{Statistical approximations on GLUE.} Mean $\pm$ std over three seeds. Diag + Mean sends variance and mean vectors. Extra uplink is per client/layer/refresh, relative to rank-8 FP32 LoRA (48 KiB/layer).}
\label{tab:rasa-statistics}
\begin{tabularx}{\textwidth}{@{}l*{5}{>{\centering\arraybackslash}X}@{}}
\toprule
& \multicolumn{3}{c}{GLUE average accuracy (\%)} & \multicolumn{2}{c}{Extra statistics uplink} \\
\cmidrule(lr){2-4}\cmidrule(l){5-6}
Statistics & FlexLoRA & ECLoRA & RB-LoRA & KiB/layer & Overhead (\%) \\
\midrule
Pure & $76.56 \pm 0.94$ & $77.64 \pm 0.22$ & $77.67 \pm 0.28$ & 0.00 & 0.00 \\
\midrule
Full & $\mathbf{78.03} \pm 0.59$ & $77.89 \pm 0.43$ & $78.04 \pm 0.56$ & 1153.50 & $+2403.13$ \\
Diag + Mean & $77.97 \pm 0.40$ & $78.11 \pm 0.24$ & $\mathbf{78.32} \pm 0.41$ & 6.00 & $+12.50$ \\
\rowcolor{rasarow}
Diag & $77.83 \pm 0.21$ & $\mathbf{78.38} \pm 0.19$ & $78.11 \pm 0.34$ & 3.00 & $+6.25$ \\
\bottomrule
\end{tabularx}
\end{table*}

\begin{table*}[t]
\centering
\caption{\textbf{Resource overhead on DomainNet-100.} ViT-B/16, six clients, shared A100s, $\lambda=0$; refresh each round. Time: three-run mean; memory: mean peak. Communication is identical across methods. Changes are relative to Pure.}
\label{tab:rasa-overhead}
\begingroup\small
\setlength{\tabcolsep}{4pt}
\renewcommand{\arraystretch}{1.08}
\begin{tabularx}{\textwidth}{@{}l*{9}{>{\raggedleft\arraybackslash}X}@{}}
\toprule
& \multicolumn{3}{c}{Round time (s/round)} & \multicolumn{3}{c}{Peak GPU memory (GiB)} & \multicolumn{3}{c}{Communication (MiB/round)} \\
\cmidrule(lr){2-4}\cmidrule(lr){5-7}\cmidrule(l){8-10}
& FlexLoRA & ECLoRA & RB-LoRA & FlexLoRA & ECLoRA & RB-LoRA & Uplink & Downlink & Total \\
\midrule
Pure & 719.6 & 752.7 & 693.4 & 3.179 & 3.175 & 3.175 & 17.510 & 17.510 & 35.020 \\
\rowcolor{rasarow}
\textbf{+ RASA} & 796.2 & 731.6 & 712.1 & 3.179 & 3.175 & 3.175 & 17.932 & 17.510 & 35.442 \\
\midrule
$\Delta$ & $+76.6$ & $-21.1$ & $+18.6$ & 0.000 & 0.000 & 0.000 & $+0.422$ & 0.000 & $+0.422$ \\
$\Delta$ (\%) & $+10.64\%$ & $-2.80\%$ & $+2.69\%$ & $0.00\%$ & $0.00\%$ & $0.00\%$ & $+2.41\%$ & $0.00\%$ & $+1.20\%$ \\
\bottomrule
\end{tabularx}\endgroup
\end{table*}

\begin{figure}[t]
\centering
\includegraphics[width=\columnwidth]{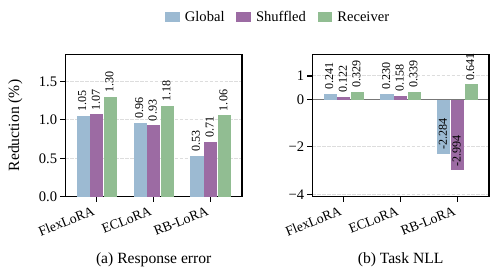}
\caption{\textbf{Effect of receiver statistics on NICO++ at rank eight.} Relative reductions from spectral readout in response error and task NLL, under the same seed. Positive values indicate improvement.}
\label{fig:receiver-mechanism}
\end{figure}

\noindent\textbf{Component contributions.}
\label{sec:ablation}
Table~\ref{tab:rasa-ablation} shows that balancing alone (w/o receiver statistics) improves over the baseline, while receiver-aware readout further boosts seen-task ROUGE-L. Update correction improves both seen and unseen ROUGE-L over the uncorrected variant, with Full RASA achieving the best scores on both splits.

\noindent\textbf{Statistical approximation.}
Table~\ref{tab:rasa-statistics} shows that Diag improves over Pure across all three methods, attaining the highest mean accuracy for ECLoRA. Its mean accuracies remain close to Full while requiring $384.5\times$ less statistics uplink (3 vs.\ 1153.5 KiB per layer per refresh). Adding mean statistics doubles the payload without consistent accuracy gains, supporting Diag as a practical accuracy--communication trade-off.

\noindent\textbf{Receiver statistics.}
\label{sec:mechanism}
Figure~\ref{fig:receiver-mechanism} shows that receiver statistics give the largest reductions in both response error and NLL across all three methods on NICO++ at rank eight. In contrast, global and shuffled statistics reduce response error but worsen NLL for RB-LoRA. This highlights the importance of matching statistics to the recipient: lower response error alone does not guarantee better task performance.

\subsection{Sensitivity and Efficiency}
\begin{figure}[t]
\centering
\includegraphics[width=\columnwidth]{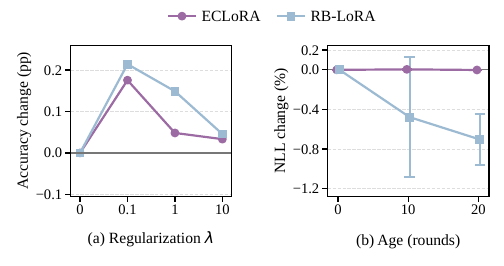}
\caption{\textbf{Sensitivity on DomainNet-100.} (a) Round-50 test accuracy change from $\lambda=0$ (seed 42). (b) Validation NLL change from fresh statistics at $\lambda=0.1$, fixing the round-50, rank-32 reference. Error bars: mean $\pm$ std over three runs.}
\label{fig:sensitivity}
\end{figure}
\label{sec:sensitivity}

\noindent\textbf{Fidelity regularization.}
Figure~\ref{fig:sensitivity}(a) evaluates round-50 test accuracy on DomainNet-100 with $\lambda\in\{0,0.1,1,10\}$ for ECLoRA and RB-LoRA. Both peak at $\lambda=0.1$, gaining approximately $0.18$ and $0.21$ accuracy points over $\lambda=0$. Gains diminish as $\lambda$ increases to 1 and 10, while remaining positive. A moderate fidelity term complements receiver adaptation; stronger regularization increasingly favors the shared spectral geometry.

\noindent\textbf{Staleness of cached activation statistics.}
We fix the round-50, rank-32 reference at $\lambda=0.1$ and replay statistics from rounds 50, 40, and 30 on the same validation examples, isolating readout sensitivity to cache age. Figure~\ref{fig:sensitivity}(b) shows nearly unchanged mean NLL for ECLoRA and reductions of approximately $0.5\%$ and $0.7\%$ for RB-LoRA at ages 10 and 20. Thus, older statistics retain useful receiver information for this fixed-state readout. This complements Eq.~\eqref{eq:regret}, which relates readout error to drift in response geometry.

\noindent\textbf{Resource overhead.}
\label{sec:overhead}
Table~\ref{tab:rasa-overhead} measures DomainNet-100 costs with six clients and statistics refreshed every round. Recorded time increases by $10.64\%$ for FlexLoRA and $2.69\%$ for RB-LoRA. ECLoRA's $-2.80\%$ difference occurs amid run-to-run variability on shared GPUs. Peak process GPU memory is unchanged at the reported precision. Statistics add $0.422$ MiB per round to the uplink, leaving downlink unchanged; total communication rises by only $1.20\%$. Less frequent refreshes amortize the additional statistic payload.

\section{Conclusion}
RASA couples response PCA with spectral-reference aggregation to reduce receiver-response loss in federated LoRA. Compact statistics guide subspace selection at fixed rank, while reference correction preserves locally learned increments. Theory establishes receiver-objective optimality and zero-local-update consistency. Experiments across vision and language tasks show gains under heterogeneity with modest communication overhead.

\clearpage
\twocolumn
\section*{LLM Usage Statement}
We used LLM-based tools, primarily GPT6-Astra and Codex, to assist with coding, polish the writing, and support self-review of the manuscript. We take full responsibility for the content.

{\small
\bibliographystyle{ieeenat_fullname}
\bibliography{main}
}
\end{document}